\documentclass[sensors,article,accept,pdftex,moreauthors]{Definitions/mdpi} 
\usepackage{color, xcolor}
\soulregister\cite7
\soulregister\ref7
\soulregister\url7
\soulregister\bf7
\soulregister\em7
\soulregister\bibitem7
\soulregister\item7

\firstpage{1} 
\pubvolume{1}
\issuenum{1}
\articlenumber{0}
\pubyear{2023}
\copyrightyear{2023}

\datereceived{ } 
\daterevised{ } 
\dateaccepted{ } 
\datepublished{ } 

\hreflink{https://doi.org/} 

\Title{A Lightweight Recurrent Grouping Attention Network for Video Super-Resolution}

\TitleCitation{A Lightweight Recurrent Grouping Attention Network for Video Super-Resolution}

\Author{Yonggui Zhu 
 $^{1}$*\orcidA{}, Guofang Li $^{2,}$\orcidB{}}

 \AuthorNames{Yonggui Zhu, Guofang Li}

\AuthorCitation{Zhu, Y.; Li, G.}

\address{%
$^{1}$ \quad  School of Data Science and Intelligent Media, Communication University of China, Beijing 100024, China; ygzhu@cuc.edu.cn 
\\
$^{2}$ \quad   School of Information and Communication Engineering, Communication University of China, Beijing 100024, China; gfli@cuc.edu.cn}

\corres{Correspondence: ygzhu@cuc.edu.cn}

\abstract{Effective aggregation of temporal information of consecutive frames is the core of achieving video super-resolution. Many scholars have utilized structures such as sliding windows and recurrent to gather spatio-temporal information of frames. However, although the performance of the constructed VSR models is improving, the size of the models is also increasing,  exacerbating the demand on the equipment. Thus, to reduce the stress on the device, we propose a novel lightweight recurrent grouping attention network. The parameters of this model are only 0.878M, which is much lower than the current mainstream model for studying video super-resolution. We design forward feature extraction module and backward feature extraction module to collect temporal information between consecutive frames from two directions. Moreover, a new grouping mechanism is proposed to efficiently collect spatio-temporal information of the reference frame and its neighboring frames. The attention supplementation  module is presented to further enhance the information gathering range of the model. The feature reconstruction module aims to aggregate information from different directions to reconstruct high-resolution features. Experiments demonstrate that our model achieves state-of-the-art performance on multiple datasets.}

\keyword{Video super-resolution; Temporal grouping attention; Attention supplementation; Feature reconstruction}

\begin{document}

\section{Introduction}
Super-resolution (SR) refers to yielding high-resolution (HR) images from the corresponding low-resolution (LR) images. As a branch of this field, video super-resolution (VSR) mainly utilizes the spatial information of the current frame and the temporal information between neighboring frames to reconstruct HR frame. At present, VSR has derived non-blind VSR \cite{Non_Blind1}, blind VSR \cite{Blind_VSR1}, online VSR \cite{Online_VSR1} and other branches \cite{CAVSR_other_fields}, and is widely used in remote sensing \cite{remote_sensing1, remote_sensing2}, video surveillance \cite{video_surveillance1, video_surveillance2}, face recognition \cite{face_recognition1, face_recognition2}, and other fields \cite{traffic_monitoring1, traffic_monitoring2}. At present, with the development of technology, the resolution of video is gradually increasing. Although these enrich our lives and facilitate tasks such as surveillance and identification, they can put more pressure on areas such as video storage and transmission. In addressing these issues, VSR technology plays an important role. However, VSR is an ill-posed problem and it is difficult to find the most appropriate reconstruction model. Thus, it remains a very worthwhile endeavor to continue to explore VSR technology.

To obtain high quality images, studies have proposed numerous effective methods. Initially, researchers utilize interpolation methods to obtain HR videos \cite{interpolation1, interpolation2}. These methods possess higher computing speed, but the results are poor. With the development of deep learning, constructing models \cite{model2, model3, Sensors2} in different domains with deep learning has become a mainstream research method. Researchers have constructed different VSR models based on deep learning and achieved a large number of results. For example, some researchers \cite{VSRSTAN, VSRVESPCN, VSREDVR, Sensors1} utilized explicit or implicit alignment to explore temporal flow between frames. This type of method can effectively align adjacent frames to the reference frame to extract high-quality temporal information. However, the alignment part increases the computational effort of the model, thus exacerbating the burden during model training and testing. Meanwhile, inaccurate optical flow often leads to errors in alignment, which affects the performance of models. Moreover, some scholars \cite{VSRTGA, VSRD3D, VSRDSMC} used 3D convolution or deformable 3D convolution to directly aggregate spatio-temporal information between different frames. Although this approach can quickly aggregate information from different times, it also incorporates a lot of temporal redundancy in features, which reduces the reconstruction ability of the model. In addition, in recent years, with the rise of Transformer, the application of Transformer to construct VSR models has also become a very popular research topic. Researchers \cite{VSRRSTT, VSRTTVSR, VSRFTVSR} applied Transformer to analyze and acquire the motion trajectory of the video to sufficiently aggregate the spatio-temporal information between consecutive frames. However, due to the relatively high amount of computation required by Transformer, this limits the further development of Transformer in the field of VSR.

In numerous studies on VSR models, although the reconstruction ability of VSR models is becoming stronger, the framework of the model is also becoming larger. Several recent classic papers VSR \cite{VSRTTVSR, VSRFTVSR, VSRETDM, VSRBasicVSR++, VSRBasicVSR, VSRFDAN} have models with parameter counts of 6M or more, which undoubtedly increase the burden of model training and testing, thus affecting the application of models in real scenarios. Thus, to ameliorate this problem, this paper focuses on the VSR model for small-scale. Our goal is to get superior VSR recovery by utilizing fewer parameters. In the specific operation, we design a model with parameters less than 1M, which achieves relatively favorable results at a scale much lower than the mainstream models. This construction of the model helped us to reduce the dependence of model on device and make it easier to apply the model to online VSRs and mobile phones in the future.

In this paper, we present a novel lightweight recurrent grouping attention network (RGAN). It is a bi-directional propagation VSR model that can effectively aggregate information from different time ranges. In RGAN, we construct the forward feature extraction module (FFEM) and the backward feature extraction module (BFEM), which are able to efficiently aggregate temporal information over long distances passed in both forward and backward directions. In addition, we propose a novel temporal grouping attention module (TGAM) which divides input frames at each time step into the reference group and the fusion group. This grouping method can fully extract the information of reference frame and adjacent frames, while ensuring the stability of the model and preventing large temporal offsets. Then, we design the attention supplementation module (ASM). This module increases the scope of information collection and can more efficiently assist the model in recovering the detailed information of frames. After utilizing FFEM and BFEM to effectively aggregate and adequately extract temporal information for different ranges, we design the feature reconstruction module (FRM) to capture features obtained from FFEM and BFEM. This module can effectively integrate the temporal information of the two propagation stages and enhance the reconstruction capability of the model. Experiments demonstrate that our model possesses better performance. The contributions of this paper are listed as follows:

$\bullet$ We design a novel lightweight recurrent grouping attention network that achieves better model performance with a small number of parameters.

$\bullet$ A new grouping method is designed to enhance the stability of model and effectively extract spatio-temporal information from reference frame and adjacent frames.

$\bullet$ We design a new attention supplement module that enhances the range of information captured by the model and facilitates the recovery of more detailed information by the model.

$\bullet$ Experiments indicate that our model achieves better results on Vid4, SPMCS, UDM10 and RED4 datasets.

The rest of the paper is organized as below. In Section 2, we describe the work related to the model. In Section 3, we introduce the specific structure of the model. In Section 4, we provide details when training and testing the model, as well as comparative results with other models and ablation studies. In section 5, we summarize the paper and present the next research plan.

\section{Related Work}

\subsection{Single Image Super Resolution}
Single image super resolution (SISR) is the basis of super-resolution. In recent years, with the development of deep learning, SR has ushered in a new revolution. Dong et al. \cite{SISRSRCNN} were the first to apply deep learning to SISR. They presented a three layers convolution neural network and achieved better effect. It was proved that deep learning possessed great potential in the field of SR. After this paper, Kim et al. \cite{SISR1} presented a very deep neural network and applied the residual network to the SR model, achieving better effect than SRCNN. Song et al. \cite{SISR2} came up with the idea of making use of additive neural network for SISR, which replaced the traditional convolution kernel multiplication operation in the calculation of output layer, saving numerous computation power. Liang et al. \cite{SISRSwimIR} introduced Swin Transformer into SISR and obtained high-quality recovered images. Tian et al. \cite{model1} proposed heterogeneous grouping blocks to enhance the internal and external interactions of different channels to obtain rich low-frequency structural information. In practice, Lee et al. \cite{Sensors3} applied the SR technique to the Satellite synthetic aperture radar, and could effectively recover the information of scatterers. Moreover, many scholars have also constructed SISR models using methods such as GAN or VAE, etc \cite{SISRGAN1, SISRTransformer1, Sensors3, Sensors4, Sensors5}. Although the SISR model can also be used to reconstruct HR video, the SISR model is only capable of capturing the spatial information of frames, and can not aggregate the temporal information between neighboring frames. As a result, the quality of the video recovered by the SISR is poor, while often suffering from artifacts and other problems. To reconstruct high-quality HR videos, researchers have shifted their focus to VSR models.

\subsection{Video Super resolution}
VSR is an extension of SISR. In VSR, the temporal information between adjacent frames play a vital role. In order to acquire perfect results, studies have built a variety of modules. For instance, Caballero et al. \cite{VSRVESPCN} applied the optical flow field which included coarse flow and fine flow to align adjacent frames, and constructed an end-to-end spatio-temporal module. Based on \cite{VSRVESPCN}, Wang et al. \cite{VSRMMCNN} combined optical flow field and long short-term memory to make more efficient use of inter-frame information and obtain more real details. Moreover, Tian et al. \cite{VSRTDAN} was the first model to substitute the deformable convolution into VSR, which amplified the feature extraction ability of the model. Based on \cite{VSRTDAN}, Wang et al. \cite{VSREDVR} proposed pyramid, cascading and deformable (PCD) module that further enhances the alignment capability of the model. Then, Xu et al. \cite{VSRTMNet} designed temporal modulation block to modulate the PCD module. Meanwhile, They conducted short-term and long-term feature fusion to better extract motion clues. These optical flow based methods have also been applied to practical work such as video surveillance, etc. Guo et al. \cite{video_surveillance2} utilized optical flow and other methods to construct the back-projection network, which can effectively reconstruct high-quality surveillance videos. Moreover, Isobe et al. \cite{VSRTGA} proposed the structure of intra-group fusion and inter-group fusion, and used 3D convolution to capture and supplement the spatio-temporal information between different groups. Ying et al. \cite{VSRD3D} proposed deformable 3D convolution with efficient spatio-temporal exploration and adaptive motion compensation capabilities. Fuoli et al. \cite{VSRRLSP} devised a hidden space propagation scheme that effectively aggregates temporal information over long distances. Based on \cite{VSRRLSP}, Isobe et al.  \cite{VSRETDM} explored the temporal differences between LR and HR space, effectively complementing the missing details in LR frames. Then, Jin et al. \cite{remote_sensing1} used the temporal difference between long and short frames to achieve information compensation for satellite VSR. Liu et al. \cite{VSRTTVSR} has designed a trajectory transformer that analyzes and utilizes motion trajectories between consecutive frames to obtain high-quality HR videos. Then, on the basis of \cite{VSRTTVSR}, Qiu et al. \cite{FTVSR} introduced the frequency domain into the VSR domain, which provided a new idea to study the VSR.

Although all of the above methods have achieved good video recovery structures, the performance of these models is achieved by utilizing larger model structures. These models can exacerbate the strain on equipment and cause significant resource loss. To avoid these problems, we propose a novel lightweight Recurrent grouping attention network, which is capable of obtaining better recovery with fewer parameters. This lightweight design can effectively reduce the loss of equipment, and possesses certain theoretical significance and practical application value.
\begin{figure*}[ht]
	\begin{center}
		\includegraphics[scale=0.73]{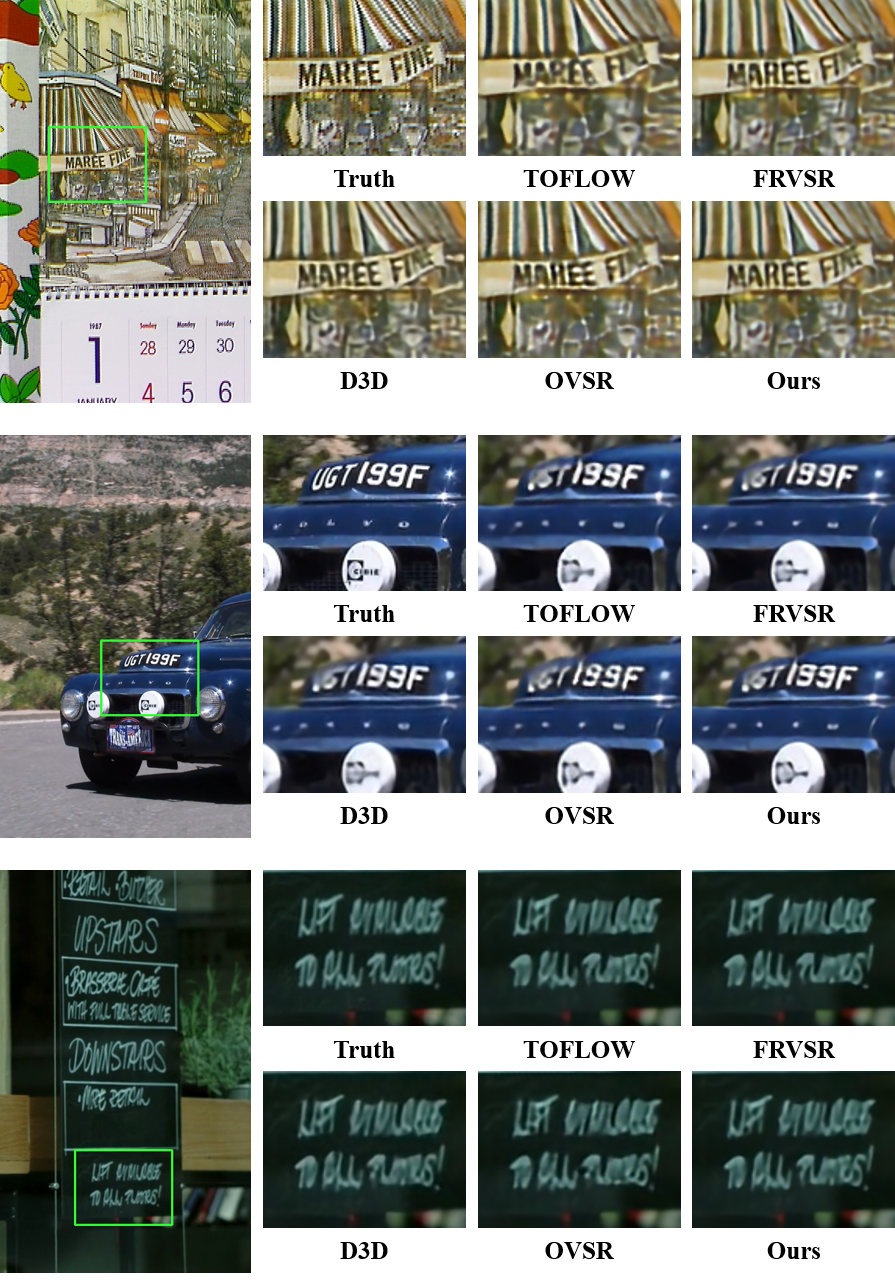}
		\caption{Qualitative comparisons on Vid4, SPMCS, and UDM10 datasets for 4$\times$VSR.}
		\label{fig:compare1}
	\end{center}
\end{figure*}

\startlandscape
\begin{figure*}[ht]
	\begin{center}
		\includegraphics[scale=0.7]{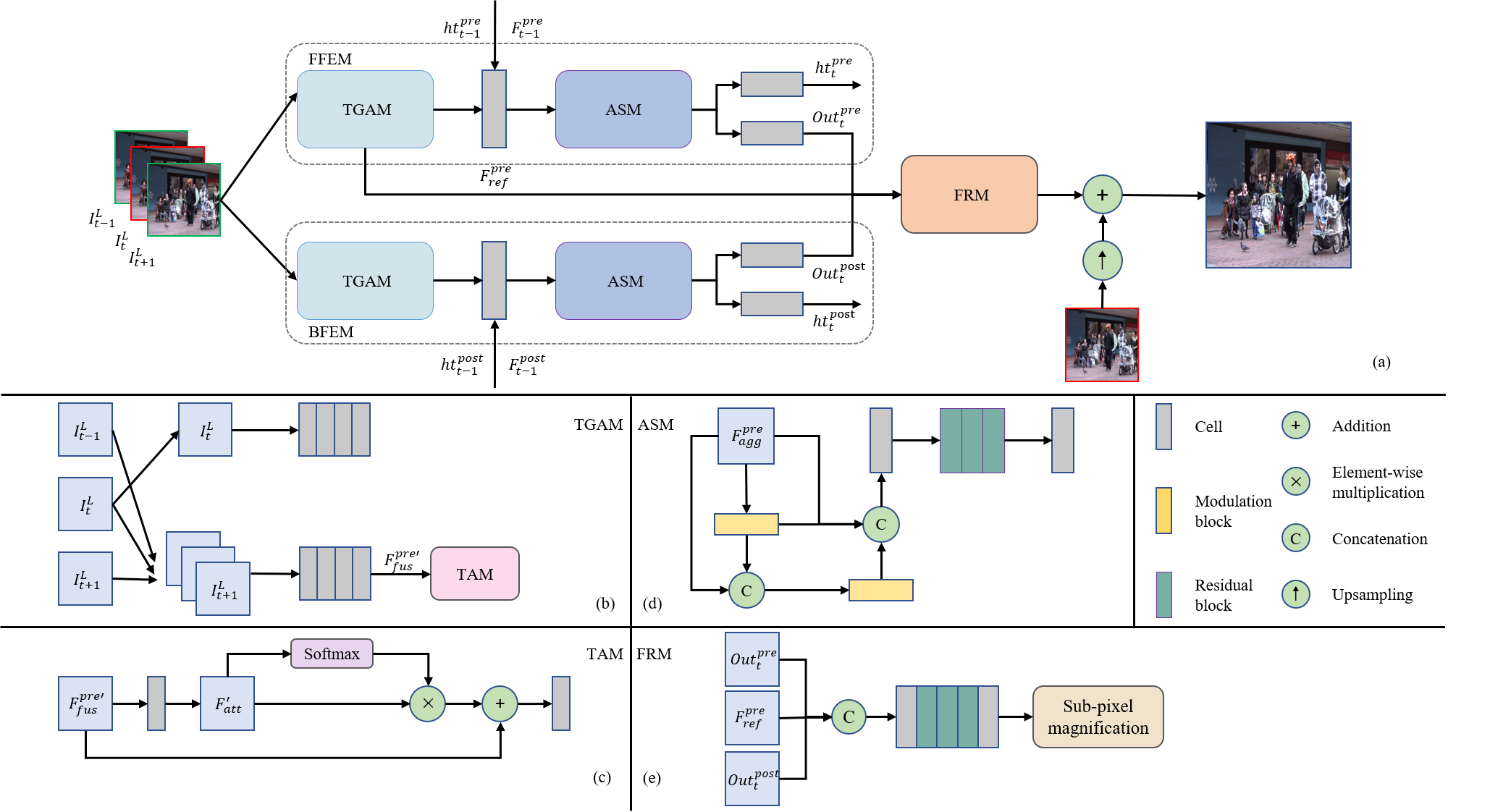}
		\caption{(a) The overall pipeline of the recurrent grouping attention network. (b) The structure of the temporal grouping attention module. (c) The structure of the temporal attention module. (d) The structure of the attention supplementation module. (e) The structure of the feature reconstruction module.}
		\label{fig:total_framework}
	\end{center}
\end{figure*}
\finishlandscape

\section{Our Method}
\subsection{Overview}
For the given consecutive frame $I^L_{0}, I^L_1, \cdots, I^L_T$, our goal is to generate the corresponding HR frame $I^H_{0}, I^H_1, \cdots, I^H_T$. Our proposed RGAN is a bi-directional propagation model where each HR frame $I^H_t$ is generated by three consecutive frames $I^L_{t-1}, I^L_t, I^L_{t+1}$, pre-hidden state $ht^{pre}_{t-1}$, post-hidden state $ht^{post}_{t+1}$, forward output feature $Out^{pre}_{t-1}$ and backward output feature $Out^{post}_{t+1}$. The structure of the model is shown in Figure \ref{fig:total_framework}(a). In the specific operation, we first input consecutive frames $I^L_{t-1}, I^L_t, I^L_{t+1}$ into FFEM and BFEM. The purpose is to gather more spatio-temporal information in different temporal directions. In FFEM and BFEM, we utilize TGAM and ASM to aggregate feature information, respectively. The role of TGAM is to perform grouping of three consecutive frames of the input and gather the grouping information. The role of ASM is to collect spatio-temporal information from another perspective and increase access to information. Then, we present the FRM module to fuse and reconstruct the outputs from FFEM and BFEM, with the aim of obtaining the final output feature. Finally, the HR frame $I^H_t$ is obtained by summing the feature generated by the model and the bicubic upsampling result of the reference frame $I^L_t$.

\subsection{Forward/Backward feature extraction module}
The FFEM and BFEM are the core of RGAN, and they have similar structures. In this section, we take FFEM as an example to introduce the specific structure of two modules. In FFEM, we first input three consecutive frames $I^L_{t-1}, I^L_t, I^L_{t+1}$ into TGAM. In TGAM, we divide the consecutive frames into the reference group and the fusion group based on the category of the extracted information and extract the information separately to obtain the forward reference group feature $F^{pre}_{ref}$ and the forward fusion group feature $F^{pre}_{fus}$. Then, we input $F^{pre}_{ref}$, $F^{post}_{fus}$, $ht^{pre}_{t-1}$ and $Out^{pre}_{t-1}$ into a cell to obtain the aggregated features $F^{pre}_{agg}$. The cell consists of a $3\times3$ convolution and a Leaky ReLU. After that, we use the ASM to further optimize the feature information of $F^{pre}_{agg}$. Finally, the optimized information is delivered into two branches. One is to utilize a cell to obtain the forward hidden state $ht^{pre}_{t}$ of the current time step, and the other is to utilize a cell to obtain the output $Out^{pre}_{t}$ of the current time step. The $ht^{pre}_{t}$ and $Out^{pre}_{t}$ will be applied to the next time step of the FFEM. Moreover, the $Out^{pre}_{t}$ is applied to the FRM to synthesize the output of the current time step. BFEM and FFEM have approximately the same structure. The difference between both modules is that after TGAM, BFEM utilizes a single cell to aggregate the information of backward reference group feature $F^{post}_{ref}$, backward fusion group feature $F^{post}_{fus}$, $ht^{post}_{t+1}$, and $Out^{post}_{t+1}$. In addition, $F^{pre}_{ref}$ in FFEM is applied to the FRM, while $F^{post}_{ref}$ in BFEM does not perform this action. In order to better represent the similar and different parts between two modules, we provide the formulas for two modules as follows:
\begin{equation}
	\begin{aligned}
		&ht^{pre}_{t}, Out^{pre}_{t} = N_{FFEM}(I^L_{t-1}, I^L_t, I^L_{t+1}, ht^{pre}_{t-1}, Out^{pre}_{t-1}), \\
		&ht^{post}_{t}, Out^{post}_{t} = N_{BFEM}(I^L_{t-1}, I^L_t, I^L_{t+1}, ht^{post}_{t+1}, Out^{post}_{t+1}),
	\end{aligned}
\end{equation}
where $N_{FFEM}$ and $N_{BFEM}$ denote the FFEM and BFEM, respectively, and $ht^{post}_{t}$ and $ Out^{post}_{t}$ indicate the hidden state and output feature obtained by the BFEM module at the current time step. The FFEM and BFEM can effectively aggregate temporal information in different directions to enhance and optimize the feature extraction capability of the model.

\begin{figure*}[ht]
	\begin{center}
		\includegraphics[scale=0.73]{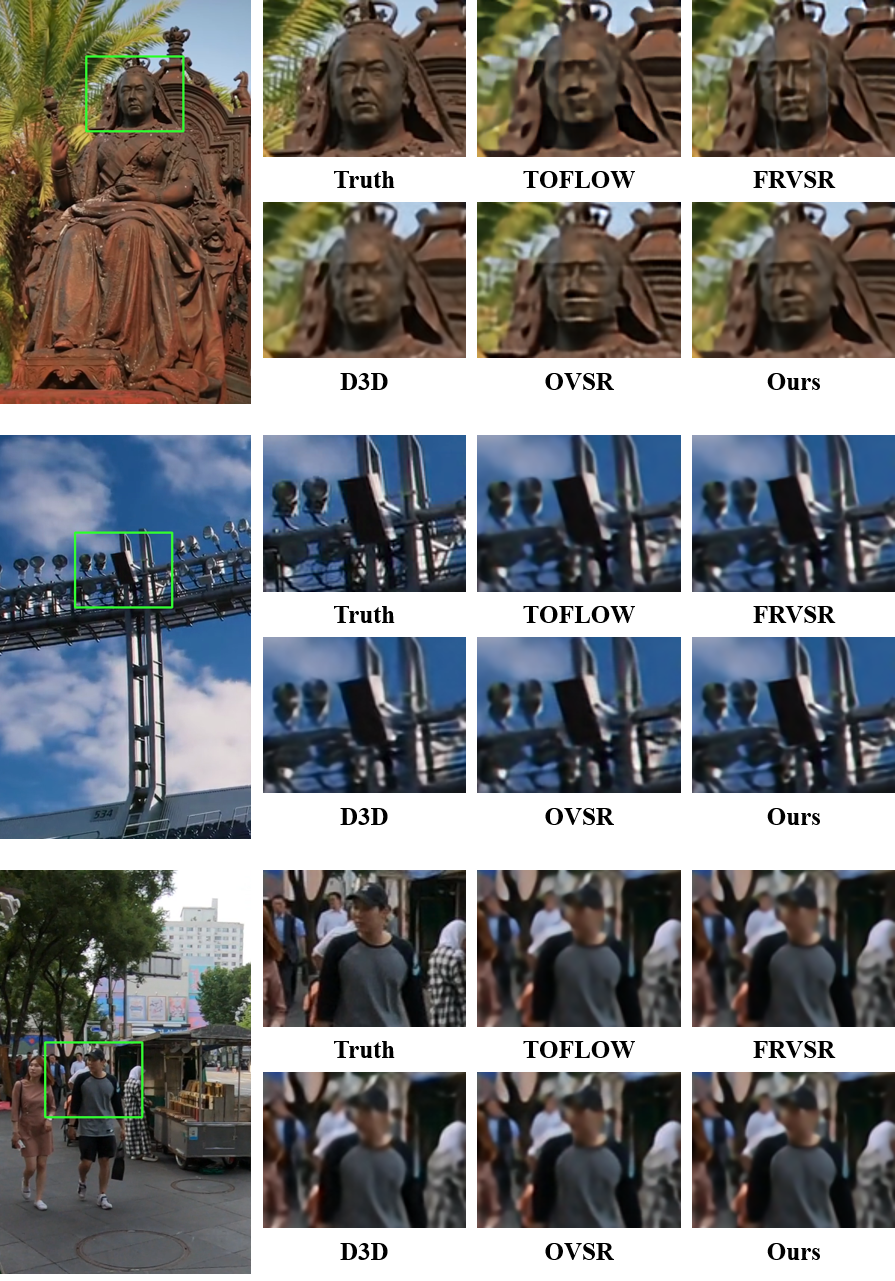}
		\caption{Qualitative comparisons on SPMCS, UDM10 and RED4 datasets for 4$\times$VSR.}
		\label{fig:compare2}
	\end{center}
\end{figure*}
\subsection{Temporal Grouping Attention Module}
The role of TGAM is to efficiently extract features between the reference frame and its neighboring frames, and it possesses the same structure in FFEM and BFEM. We present the ASM in FFEM as a case study to introduce the specific structure of the ASM. The structure of TGAM is shown in Figure \ref{fig:total_framework}(b). For the input three consecutive frames $I^L_{t-1}, I^L_t, I^L_{t+1}$, we first divide them into the reference group $I^L_t$ and the fusion group $I^L_{t-1}, I^L_t, I^L_{t+1}$. The purpose of the reference group is to maintain the temporal stability of model and prevent large shifts in generated features. Meanwhile, the aim of the fusion group is to efficiently aggregate temporal information between adjacent frames. For the reference group, we apply four cells to obtain feature $F^{pre}_{ref}$ of the reference group. The formula for this operation is shown as follows:
\begin{equation}
	F^{pre}_{ref} = \sum^{4}_{i=1}C^i(I^L_t, \theta^i_{ref}),
\end{equation}
where $C^i$ and $\theta^i_{ref}$ represent cells and the corresponding parameters. The cells below are expressed in the same manner. The role of the reference group is to exclude temporal interference and extract feature information only from the spatial domain, preventing large shifts in the FFEM during iterations. For the fusion group, we employ four cells to obtain feature $F^{pre'}_{fus}$. Then, we use the temporal attention module (TAM) to further collect inter-frame information. The TAM is inspired by \cite{VSRTGA}. Firstly, we apply a cell to obtain the feature $F^{'}_{att}$. Later, we select one of the channel feature maps in $F^{'}_{att}$ and compute the attention mapping using the softmax function in the depth dimension. Then, we utilize element-wise multiplication to multiply the remaining channels in $F^{'}_{att}$ and temporal weights to obtain the attention feature $F^{pre}_{att}$. The role of TAM is to add a new path for extracting features of continuous frames, thus better extracting detailed spatio-temporal information of continuous frames. Finally, we splice $F^{pre'}_{fus}$ and $F^{pre}_{att}$ in the channel dimension and perform extraction using a single cell to obtain the final fusion group feature $F^{pre}_{fus}$. The structure of the TAM is illustrated in Figure \ref{fig:total_framework}(c). The formula for this operation is shown as follows:
\begin{equation}
	F^{pre}_{ref} = N_{TAM}\left(\sum^{4}_{i=1}C^i(I^L_t, \theta^i_{fus})\right),
\end{equation}
where $N_{TAM}$ and $\theta^i_{fus}$ denote the TAM and cell parameters, respectively. The role of the fusion group is to initially extract the spatio-temporal features of consecutive frames and prepare for the next feature supplementation and reconstruction.

\subsection{Attention Supplementation Module}
After fusing the information from different stages in FFEM and BFEM, we apply ASM to further increase the information acquisition range of the model to supplement the missing detail information. The core of ASM is to use the densely connected structure and spatial attention mechanism to obtain more spatio-temporal information, and it possesses the same structure in FFEM and BFEM. We describe the specific structure of the ASM using the ASM in FFEM as an example. The structure of ASM is displayed in Figure \ref{fig:total_framework}(d). Firstly, we construct a modulation block that consists of a $1\times1$ convolution, a Leaky ReLU, a $3\times3$ convolution, a Leaky ReLU and a spatial attention module \cite{SISRRCAN}. Then, we apply a modulation block to extract $F^{pre}_{agg}$ to obtain the feature $F^1_{ASM}$. Next, we splice $F^{pre}_{agg}$ and $F^1_{ASM}$ in the channel dimension and then extract the model with a modulation block to obtain the output feature $F^2_{ASM}$. After that, we splice $F^{pre}_{agg}$, $F^1_{ASM}$ and $F^2_{ASM}$ in the channel dimension and then aggregate three different groups of features with a single cell. Finally, we employ three residual blocks \cite{SISRESDNResblock} to further optimize obtained features. The ASM is able to further extract the information that has been fused and increase the range of information accessed by the model. In addition, the densely connected design structure can effectively aggregate different types of information and enhance the utilization of features.

\subsection{Feature Reconstruction Module}
After FFEM and BFEM, we use FRM to aggregate two groups of temporal information obtained from different directions. The structure of FRM is displayed in Figure \ref{fig:total_framework}(e). Firstly, we utilize a cell to fuse $F^{pre}_{out}$, $F^{post}_{out}$, and $F^{pre}_{ref}$. The purpose of applying $F^{pre}_{ref}$ to this aggregation part is to further enhance the stability of the model and prevent bias in the generated feature. Afterwards, we apply three residual blocks to further optimize the fused features and employ a $3\times3$ convolution to adjust the number of channels of the output feature to 48 to facilitate upsampling. Finally, we use sub-pixel magnification \cite{SISRESPCN} to obtain the final HR features. The FRM can effectively aggregate the information generated by FFEM and BFEM in different directions, possesses strong reconstruction capability, and can obtain high-quality HR features.

\startlandscape
\begin{table}[htbp]
	\centering
	\caption{\small{Quantitative comparison (PSNR(dB) and SSIM) on Vid4, SPMCS11, UDM10 and RED4 for 4$\times$VSR. The bolded portion indicates the best performance. The runtime is calculated on the LR image of $320\times180$.}}
	\begin{tabular}{c|c|c|c|c|c|c}
		\hline
		Method & Bicubic & TOFLOW \cite{VSRTOFLOWDatasetsVimeo90K} & FRVSR \cite{VSRFRVSR} & D3D \cite{VSRD3D} & OVSR \cite{VSROVSR} & Ours\\
		\hline
		Params(M) & -/- & 1.4 & 5.1   & 2.6 & 0.895 & 0.878 \\
		\hline
		Runtime(ms) & -/- & 493 & 114 & 119 & 19 & 18 \\
		\hline
		Vid4 & 21.80/0.5246 & 25.85/0.7659 & 26.69/0.8103 & 26.72/0.8134 & 26.26/0.7984 & \textbf{26.80/0.8149} \\
		\hline
		SPMCS & 23.29/0.6385  & 27.86/0.8237 & 28.16/0.8421 & 28.71/0.8515 & 27.79/0.8433 & \textbf{28.95/0.8608} \\
		\hline
		UDM10 & 28.47/0.8253  & 36.26/0.9438 & 37.09/0.9522 & 37.36/0.9545 & 36.80/0.9511 & \textbf{37.93/0.9575} \\
		\hline
		RED4 & 26.14/0.7292 & 27.93/0.7997 & 29.71/0.8356 & 29.50/0.8319 & 29.45/0.8285 & \textbf{29.82/0.8383} \\
		\hline
	\end{tabular}
	\label{compare}
\end{table}

\begin{table}[htbp]
	\centering
	\caption{Quantitative comparison of ablation study of temporal grouping attention module. 'RG' represents the reference group.}
	\setlength{\tabcolsep}{3mm}
	\begin{tabular}{c|c|c|c|c|c|c|c}
		\hline
		Method & RG & TAM & Param(M) & Vid4 & SPMCS & UDM10 & RED4 \\
		\hline
		1 &  &  & 0.771 & 26.35/0.8027 & 28.21/0.8454 & 37.12/0.9536 & 29.62/0.8338 \\
		\hline
		2 &  & $\checkmark$ & 0.815 & 26.39/0.7997 & 27.99/0.8354 & 37.32/0.9534 & 29.64/0.8339 \\
		\hline
		3 & $\checkmark$ &  & 0.834 &  26.53/0.8091 & 28.16/0.8477 & 37.40/0.9549 & 29.67/0.8341 \\
		\hline
		4 & $\checkmark$ & $\checkmark$ & 0.878 & \textbf{26.80/0.8149} & \textbf{28.95/0.8608} & \textbf{37.93/0.9575} & \textbf{29.82/0.8383} \\
		\hline
	\end{tabular}
	\label{compare_TGAM}
\end{table}
\finishlandscape

\section{Experiments}

\subsection{Training Datasets and Details}
\textbf{Datasets} In this paper, we utilize Vimeo-90K \cite{VSRTOFLOWDatasetsVimeo90K} as the training dataset. This dataset contains over 90K video sequences, each consisting of seven consecutive frames with a resolution of $448\times256$. Moreover, we apply Vid4 \cite{DatasetsVid4}, SPMCS \cite{DatasetsSPMCS11}, UDM10 \cite{VSRPFNLDatasetsUDM10} and RED4 \cite{DatasetsRED4} as test datasets. These datasets contain sequences of different lengths and resolutions of natural environments, human landscapes and other types of sequences, which can effectively indicate the performance and generalization ability of the model. Then, we utilize peak signal-to-noise ratio (PSNR) and structural similarity (SSIM) as evaluation metrics and perform tests on the Y channel in YCbCr space.

\textbf{Implementation details} For training, we choose the HR frames with the size of $256\times256$, which are randomly selected among the sequences in the Vimeo-90K dataset and each sequence selects the same region. The size of LR image is $64\times64$ by applying Gaussian  kernel with the standard deviation of $\sigma=1.6$ and $4\times$ downsampling. The initial learning rate is set to $1\times10^{-4}$ and decayed with a decay factor of 0.5 every 25 epochs until 75 epochs. The batch size is 8. Moreover, to ensure that all video sequences can be adequately trained and tested, we copy the first and last frames to complement the missing adjacent frames of the first and last frames. During training and ablation studies, we input 7 consecutive frames for training. Meanwhile, to increase the range of the training dataset, we perform random rotations and flips of the input sequence. During testing, the number of frames input at a time depends on the length of the sequence. All training and assessments are experimented with Python 3.8, PyTorch 1.8 and RTX 3090 GPUs.

\subsection{Comparison with State-of-the-art Methods}
In this section, we compare our model with the state-of-the-art model. we compare our model with several state-of-the-art models. The models for comparison include TOFLOW \cite{VSRTOFLOWDatasetsVimeo90K}, FRVSR \cite{VSRFRVSR}, D3D \cite{VSRD3D} and OVSR \cite{VSROVSR}. TOFLOW explored relationships between neighboring frames using task-oriented motion cues. FRVSR adopted the HR result of the previous frame to generate the current frame, constructing the uni-directional VSR model. D3D designed deformable 3D convolution to directly aggregate spatio-temporal information in continuous frames. OVSR devised a bi-directional omniscient network that effectively aggregates past, present and future temporal information.

It is well known that different training datasets and downsampling methods affect model performance. Thus, to ensure the fairness of comparison, we retrain these models using the same training set and Gaussian downsampling. Moreover, to better compare at the same parameter magnitude, we adjusted the number of channels and depth of the OVSR. The quantitative comparison results are recorded in Table \ref{compare}. Comparing with TOFLOW, FRVSR and D3D, our model is far superior to these models in terms of performance and runtime. These indicate that the small scale models we designed have outperformed these classical models in terms of performance. Moreover, compared with OVSR, we can notice that with similar parameters and runtime, our model is far better than OVSR in terms of performance. These demonstrate that our model has state-of-the-art performance with small-scale parameters.

After making quantitative comparisons, we also made qualitative comparisons of these models. The qualitative comparison results are displayed in Figure \ref{fig:compare1} and Figure \ref{fig:compare2}. In Figure \ref{fig:compare1}, we can notice that our model offers better recovery in terms of numbers, etc. Moreover, in Figure \ref{fig:compare2}, we can demonstrate that our model possesses better ability for detail and edge restoration. These further indicate that our proposed model has state-of-the-art performance.

\begin{figure*}[ht]
	\begin{center}
		\includegraphics[scale=0.9]{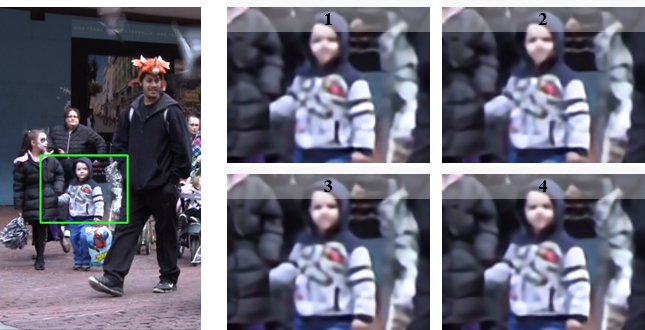}
		\caption{Qualitative comparison results of temporal grouping attention module for 4$\times$VSR.}
		\label{fig:ablation1}
	\end{center}
\end{figure*}

\subsection{Ablation Studies}
\textbf{Ablation studies of temporal grouping attention module} In the processing of three consecutive frames $I^L_{t-1}, I^L_t, I^L_{t+1}$, we propose two innovations. One is to process the reference frame as a separate group, and the other is to design the TAM in extracting features from three consecutive frames. To demonstrate the validity of these two constructions, we design corresponding ablation experiments. The ablation experiments consist of three configurations, which are removing the reference frame and TAM, only removing the reference group and only removing the TAM. The quantitative comparison results of the model are summarized in Table \ref{compare_TGAM}. In Table \ref{compare_TGAM}, we can find that when the reference group is removed, the performance of the model does not change much, with or without the addition of the TAM module. This suggests that it is meaningful to group reference frames individually. Then, with the addition of the reference group, supplementing the TAM effectively enhances the performance of the model, which proves that supplementing the TAM in the model is a worthwhile endeavor. Finally, comprehensive comparison indicates that with the addition of the reference group and the TAM, the model shows a significant improvement in performance with a small increase in parameters. This suggests that the addition of the reference group and the TAM module has positive implications for the model. Moreover, the qualitative comparison results of four groups of models are displayed in Figure \ref{fig:ablation1}. The comparison indicates that with the addition of the reference group and TAM, the model has a better recovery ability.

\begin{table}[htp]
	\centering
	\caption{Quantitative comparisons of ablation study of temporal grouping attention module.}
	\begin{tabular}{c|c|c|c}
		\hline
		& RGAN-N & RGAN-S & RGAN\\
		\hline
		Param(M) & 0.937 & 0.877 & 0.878 \\
		\hline
		Vid4 & 26.72/0.8124 & 26.40/0.800 & \textbf{26.80/0.8149} \\
		\hline
		SPMCS &  28.84/0.8581 & 28.30/0.8419 & \textbf{28.95/0.8608} \\
		\hline
		SPMCS &  37.77/0.9563 & 37.06/0.9520 & \textbf{37.93/0.9575} \\
		\hline
		SPMCS &  29.72/0.8358 & 29.57/0.8312 & \textbf{29.82/0.8383} \\
		\hline
	\end{tabular}
	\label{compare_ASM}
\end{table}

\textbf{Ablation studies of attention supplementation module} In FFEM and BFEM, after fusing different types of spatio-temporal features, we design the ASM module to further enhance the information extraction range of the model and supplement the missing temporal information. To demonstrate the role of ASM, we construct the ablation experiment RGAN-N by replacing the ASM module with the residual block. Moreover, in the modulation block in ASM, we supplement the spatial attention module to further enhance the model's performance. To demonstrate the importance of this module, we designed the model without the spatial attention module, named RGAN-S. The quantitative comparison results of these ablation experiments are displayed in Table \ref{compare_ASM}. Comparing RGAN and RGAN-N, we can indicate that the ASM module has a positive effect on the performance of the model. Moreover, comparing RGAN-S and RGAN, it can be seen that adding the spatial attention module can effectively improve the performance of the model. These prove that the ASM module we designed is efficient and meaningful.

\section{Conclusion}
In this paper, we propose a novel lightweight recurrent grouping attention network which centers on obtaining better VSR results with small scale parameters. We design forward feature extraction module and backward feature extraction module to obtain sufficient temporal information from two directions. The temporal grouping attention module is proposed to efficiently aggregate temporal information between reference frame and adjacent frames. Moreover, the attention supplement module is used to further optimize the fused information and expand the information collection range of the model. Finally, we apply the feature reconstruction module to efficiently aggregate and restructure the information in different directions to obtain high-quality HR features. Experiments demonstrate that our models achieve excellent performance. The scale of our model is much lower than the current mainstream video super-resolution model, which means that our model is more suitable for applications in remote sensing and video surveillance, etc. In the following research, we will build the VSR model for satellite video and virtual reality based on this model. Moreover, we will further optimize the model in the areas of attention module, generation module, and loss function to improve its performance.

\authorcontributions{Conceptualization, Y.Z.; methodology, Y.Z.; software, G.L.; validation, G.L.; formal analysis, Y.Z.; investigation, G.L.; resources, Y.Z.; data curation, G.L.; writing---original draft preparation, G.L.; writing---review and editing, Y.Z.; visualization, G.L.; supervision, Y.Z.; project administration, Y.Z.; funding acquisition, Y.Z. All authors have read and agreed to the published version of the manuscript.}

\funding{Supported by the National Natural Science Foundation of China(No. 11571325) and the Fundamental Research Funds for the Central Universities(No. CUC2019 A002).}

\institutionalreview{\hl{}} 

\informedconsent{\hl{}}


\dataavailability{Our code is available at \url{https://github.com/karlygzhu/RGAN}.}

\conflictsofinterest{The authors declare no conflict of interest. \hl{} 
}


\begin{adjustwidth}{-\extralength}{0cm}
\reftitle{References}

\PublishersNote{}
\end{adjustwidth}

\end{document}